%% file: main.tex
\documentclass[10pt,twocolumn,letterpaper]{article}
\usepackage{multirow}
\usepackage[misc]{ifsym}
\usepackage[pagenumbers]{cvpr} 

\input{preamble}

\definecolor{cvprblue}{rgb}{0.21,0.49,0.74}
\usepackage[pagebackref,breaklinks,colorlinks,citecolor=cvprblue]{hyperref}
\usepackage{amssymb}
\usepackage{subcaption}
\usepackage{caption}
\usepackage{cuted}
\usepackage{algorithmic,algorithm}

\newcommand\blfootnote[1]{%
  \begingroup
  \renewcommand\thefootnote{}\footnote{#1}%
  \addtocounter{footnote}{-1}%
  \endgroup
}
\usepackage[accsupp]{axessibility}

\def\confName{CVPR}
\def\confYear{2024}

\begin{document}

\title{Diversified and Personalized Multi-rater Medical Image Segmentation}

\author{Yicheng Wu\textsuperscript{1(\Letter),*}, Xiangde Luo$^{2,*}$,
Zhe Xu$^{3}$, 
Xiaoqing Guo$^{4}$, 
Lie Ju$^{1}$,
Zongyuan Ge$^{1}$, \\
Wenjun Liao$^{2}$,
Jianfei Cai$^{1}$
\\
$^{1}$Monash University, $^{2}$University of Electronic Science and Technology of China, \\ 
$^{3}$The Chinese University of Hong Kong, $^{4}$University of Oxford\\
}

\maketitle

\blfootnote{$^*$ Equal Contribution. Correspondence to yicheng.wu@monash.edu. This research is supported by the Monash FIT Start-up Grant.}

\begin{abstract}
Annotation ambiguity due to inherent data uncertainties such as blurred boundaries in medical scans and different observer expertise and preferences has become a major obstacle for training deep-learning based medical image segmentation models. To address it, the common practice is to gather multiple annotations from different experts, leading to the setting of \textbf{multi-rater medical image segmentation}. Existing works aim to either merge different annotations into the ``groundtruth'' that is often unattainable in numerous medical contexts, or generate diverse results, or produce personalized results corresponding to individual expert raters. Here, we bring up a more ambitious goal for multi-rater medical image segmentation, i.e., obtaining both diversified and personalized results. 
Specifically, we propose a two-stage framework named \textbf{D-Persona} (first \textbf{D}iversification and then \textbf{Persona}lization).
In Stage \uppercase\expandafter{\romannumeral1}, we exploit multiple given annotations to train a Probabilistic U-Net model, with a bound-constrained loss to improve the prediction diversity. In this way, a common latent space is constructed in Stage \uppercase\expandafter{\romannumeral1}, where different latent codes denote diversified expert opinions.
Then, in Stage \uppercase\expandafter{\romannumeral2}, we design multiple attention-based projection heads to adaptively query the corresponding expert prompts from the shared latent space, and then perform the personalized medical image segmentation.
We evaluated the proposed model on our in-house Nasopharyngeal Carcinoma dataset and the public lung nodule dataset (i.e., LIDC-IDRI). Extensive experiments demonstrated our D-Persona can provide diversified and personalized results at the same time, achieving new SOTA performance for multi-rater medical image segmentation. Our code will be released at \url{https://github.com/ycwu1997/D-Persona}.
\end{abstract}

\section{Introduction}
\label{sec:intro}
To build up a powerful computer-aided diagnosis (CAD) system, automatic medical image segmentation is an indispensable step in providing quantitative measurements for interested medical targets \cite{isensee2021nnu}. Despite deep-learning based medical image segmentation methods have achieved great progress \cite{falk2019u,hatamizadeh2022unetr,wu2022mutual}, their deployment in clinical settings \cite{bernard2018deep} remains skeptical since it is still hard for them to accurately segment challenging targets such as the primary gross tumor volume (GTVp) \cite{marin2022deep,luo2023deep}, blurred lesions \cite{carass2017longitudinal,wu2023coact} and complex anatomical structures \cite{li2017deep,fu2020rapid}.

\begin{figure*}[htp]
  \centering
   \includegraphics[width=1\linewidth]{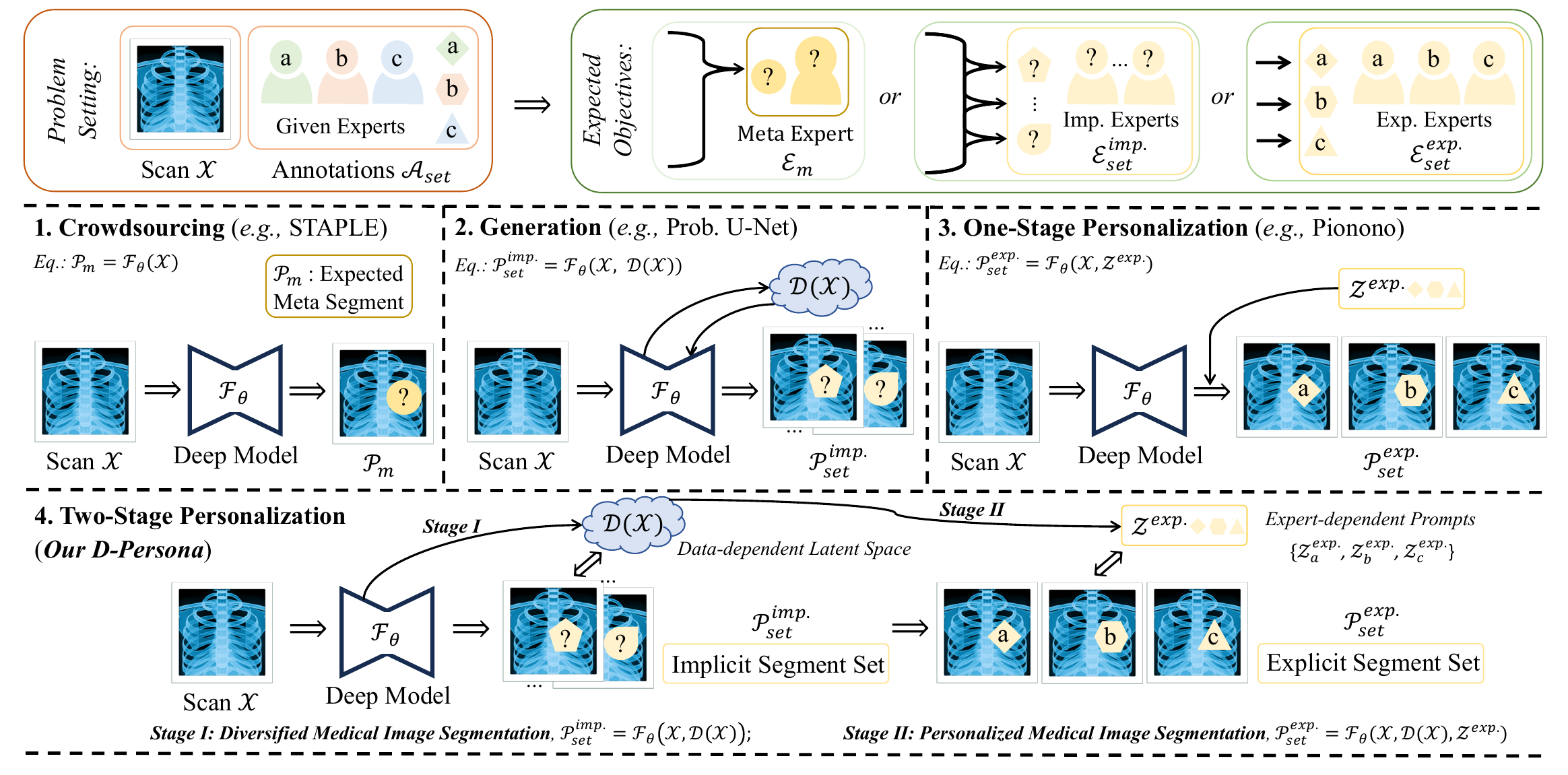}
   \caption{Overview of scheme designs in multi-rater medical image segmentation. Top: problem setting and expected objectives (\ie, meta, implicit, or explicit experts); Middle: existing methods including crowdsourcing, generation, and one-stage personalization; Bottom: our proposed two-stage framework, providing both diversified and specifically personalized segmentation simultaneously.}
   \label{fig_overview}
\end{figure*}
One key reason lies in the ``\textbf{\textit{annotation ambiguity}}'', which mainly comes from two aspects: \textit{data-level uncertainties} and \textit{observer-level preferences}.
The first aspect refers to the inherent uncertainties in the data such as irregular shapes and varied locations of a medical target (\eg, MS lesions \cite{carass2017longitudinal}) due to its great variability, and blurred boundaries around the target regions caused by limited imaging resolutions in medical scans.
These inherent data uncertainties unavoidably affect the process of expert annotations \cite{menze2014multimodal}, resulting in inconsistent training labels that are likely to confound data-driven segmentation models.
The second aspect refers to the diversity in domain expertise and personal preferences among annotators, which leads to a range of interpretations for a single target. In other words, there is no ``{absolutely correct label}'', which we call \textbf{\textit{meta segment}} in this work. It has been shown in numerous medical contexts that the meta segment is unattainable~\cite{lin2019deep}.

To mitigate the ``{\textit{annotation ambiguity}}'', a standard practice in medical settings is to gather multiple annotations from different observers \cite{marin2022deep,rahman2023ambiguous}, leading to multi-rater medical image segmentation \cite{ji2021learning} (see Fig.~\ref{fig_overview} Top-left). The existing works dealing with medical multi-rater annotations can be generally categorized into three groups: crowdsourcing methods, generation methods, and personalization methods, each aligning with a different objective: becoming meta expert, implicit experts, or explicit experts (Fig.~\ref{fig_overview} Top-right). In particular, the crowdsourcing methods \cite{warfield2004simultaneous} (Fig.~\ref{fig_overview} Middle-left) assume there exists one and only one meta segment and combining multiple annotations can approach it. The generation methods \cite{kohl2018probabilistic,rahman2023ambiguous} (Fig.~\ref{fig_overview} Middle) aim to generate diversified outputs, providing varying and plausible expert opinions as in the specialist consultation or case conference. Their generated segment set is implicit and unordered. The personalization methods \cite{HumanError2020,liao2023transformer,schmidt2023probabilistic} (Fig.~\ref{fig_overview} Middle-right) are recent ones, aiming to generate the explicit segment set, \ie, the corresponding expert results, which enables the tracking and analysis of individual raters' performance and preferences, so as to enhance the efficacy of quality control and clinical training.

Given the interrelated nature and distinct values of both implicit and explicit segment sets in the medical domain \cite{marin2022deep,van2021interobserver}, we are prompted to ask: ``\textit{Can we achieve the goals of both generation and personalization methods, i.e., producing both diversified and personalized results in multi-rater medical image segmentation?}''
To achieve this goal, in this paper, we propose a two-stage framework \textbf{D-Persona}, as shown in Fig.~\ref{fig_overview} Bottom. Our basic idea is to divide the task into two stages. Stage \uppercase\expandafter{\romannumeral1} focuses on learning a common latent space to capture the annotation variability, which can be drawn to generate diversified results as potential expert opinions. On the other hand, Stage \uppercase\expandafter{\romannumeral2} focuses on generating personalized segmentation results by reusing the learned latent space in Stage \uppercase\expandafter{\romannumeral1} and extracting the corresponding expert prompts from it.

Our major contributions can be summarized as follows.
\begin{itemize}
\item We set a new grand goal for multi-rater medical image segmentation, \ie, aiming to generate diverse expert opinions and personalized results simultaneously. We propose a novel unified two-stage framework to address the diversity and personalization requirements, respectively.

\item In Stage \uppercase\expandafter{\romannumeral1}, we propose a bound-constrained training loss, which predicts the intersections and unions of annotations and introduces flexibility in uncertain areas. This substantially enhances segmentation diversity and increases the variety in the latent space.
\item In Stage \uppercase\expandafter{\romannumeral2}, we introduce an attention-based projection mechanism for personalized medical image segmentation, which can adaptively query expert prompts from the shared latent space. 
\item 
Extensive experiments on two datasets demonstrate our proposed framework can provide diversified and personalized segmentation results simultaneously, setting new state-of-the-art (SOTA) performance for multi-rater medical image segmentation.
\end{itemize}

\section{Related work}
\label{sec:related_work}
In this section, we review existing multi-rater medical image segmentation methods as well as noisy learning works that are related to annotation ambiguity.

\subsection{Crowdsourcing-based Segmentation}
Crowdsourcing \cite{orting2019survey} is a common solution to reduce the expensive annotation costs by eliciting imperfect labels from crowds. In medical scenarios, crowdsourced labels can be collected from junior medical students and are usually quite noisy. To facilitate the model training, crowdsourcing methods \cite{valentini2016international,lee2018international} typically adopt several label-fusing strategies to achieve consensus and reduce noise, \eg, the conventional majority-voting strategy. However, this intuitive way ignores the annotators' differences either in domain expertise or personal preferences. Therefore, the Simultaneous Truth and Performance Level Estimation (STAPLE) \cite{warfield2004simultaneous} was further proposed to utilize the annotators' performance as weights to fuse crowdsourced labels, which is a more convincing way. It has been widely used in many medical tasks \cite{mahbod2021cryonuseg} to produce the ``golden standard''.

Nevertheless, in numerous medical contexts, there is no such golden standard and the meta segmentation remains elusive \cite{lee2018international} as in the case of challenging  GTVp delineation. Thus, recent multi-rater image segmentation efforts have shifted the focus to generate either diversified implicit or personalized explicit segments, instead of the meta segment.
Note that, there are also other methods, not aiming for explicit meta segments. For example,  MAX-MIG \cite{cao2018max} proposed to maximize the mutual information between the input data and crowdsourced labels for the model training. 
To better represent the expert disagreements, randomly selecting a label for training rather than using consensus could achieve better model calibration performance \cite{jungo2018effect,jensen2019improving}.

\subsection{Generation-based Segmentation}
Various generative techniques have emerged for medical tasks \cite{celard2023survey}. For example, \cite{wang2023quantitative} used generative adversarial networks (GANs) to synthesize missing brain modalities, significantly improving the diagnosis performance. In terms of segmentation, Probabilistic U-Net \cite{kohl2018probabilistic} proposed a conditional Variational Autoencoder (cVAE) for ambiguous medical image segmentation, which learns a latent space to represent the segmentation ambiguity. To increase the segmentation diversity, multi-scale improvements \cite{baumgartner2019phiseg,kohl2019hierarchical} were proposed to enhance the original architecture. Moreover, \cite{rahman2023ambiguous} used a powerful diffusion-based model for ambiguous medical image segmentation, achieving superior performance.

Despite the progress, these methods mainly focus on generating diversified implicit segmentation results and they are unable to achieve personalized segmentation. In contrast, our work focuses on generating both diversified and personalized segments in a unified way.

\subsection{One-stage Personalization Segmentation}
Personalized segmentation directly incorporates expert preferences into the model training process \cite{HumanError2020}. For example, \cite{ji2021learning} used an explicit expertness code to denote the modified training labels. \cite{liao2023transformer} employed learnable queries to represent corresponding expert annotations. Furthermore, \cite{schmidt2023probabilistic} constructed several normal distributions with learnable parameters to represent the inter- and intra-observer variability. 
All these one-stage works strive for a one-to-one correspondence in multi-rater medical image segmentation. Conversely, our proposed D-Persona model not only delivers improved personalized results but also captures a variety of expert segmentation opinions.

\subsection{Noisy Learning}
\label{related_work_noise}
Label noise is highly related to annotation ambiguity \cite{tanno2019learning}. The common practice for performing noisy learning can be divided into two categories: (1) detecting label noise and then rectifying the training samples; and (2) developing a noise-robust framework from noisy distribution. For example, in the first category, the noise transition matrix \cite{liu2015classification,menon2015learning,patrini2017making,cheng2022instance} can be estimated and used to build a statistically consistent classifier with the inferred clean distribution. Also, sample selection works \cite{li2019dividemix,wei2022self,patel2023adaptive,ju2022improving} were proposed to design effective selection strategies to detect and reduce the impact of noisy samples during the training. 
For the second category, noise-robust frameworks are developed with a variety of robust loss functions \cite{ghosh2017robust,zhang2018generalized,wang2019symmetric} and modified architecture designs~\cite{sukhbaatar2014training,zhong2019graph}, which focus on the adaptive training from the original distribution.

In contrast, similar to the recent efforts \cite{liao2023transformer,schmidt2023probabilistic} in multi-rater medical image segmentation, we do not consider label noise here. Instead, we consider label diversity caused by different expert preferences and inherent data uncertainty, aiming to model the diversity and personalize the results.

\section{Methods}
\label{sec:method}
\begin{figure*}[htp]
  \centering
   \includegraphics[width=1\linewidth]{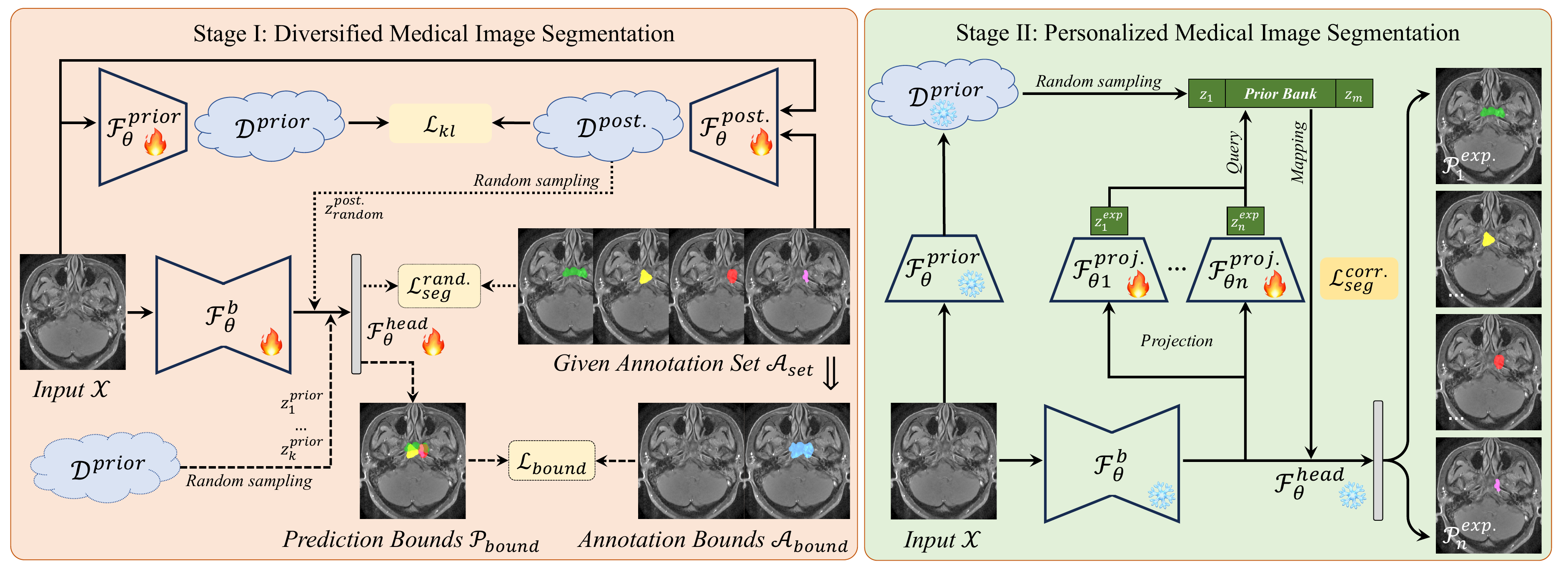}
   \caption{Pipeline of our proposed D-Persona framework for multi-rater medical image segmentation. Left: \textit{Stage \uppercase\expandafter{\romannumeral1}} is designed to construct a common latent space where different latent codes lead to diversified segmentation results; Right: \textit{Stage \uppercase\expandafter{\romannumeral2}} performs the personalized segmentation by individual projection heads to mimic the corresponding expert raters.}
   \label{fig_pipeline}
\end{figure*}
Fig.~\ref{fig_pipeline} shows the overall pipeline of our proposed D-Persona framework for multi-rater medical image segmentation, which consists of two major stages. Stage~\uppercase\expandafter{\romannumeral1} aims to learn a common latent space from $n$ experts' annotations $A_{set} = \{A_1, \cdots, A_n\}$, which can then be sampled to generate diverse segmentation results. We use the Probabilistic U-Net \cite{kohl2018probabilistic} as the baseline, including a U-Net backbone $F^b_{\theta}$ for feature extraction, a prediction head $F^{head}_{\theta}$ for the mapping from the latent space to segmentation results, and two separate encoders $F^{prior}_{\theta}$ and $F^{post.}_{\theta}$ for the prior and posterior distribution generations, respectively. Stage~\uppercase\expandafter{\romannumeral2} aims to find specific expert prompts in the latent space corresponding to the $n$ experts for personalized segmentation, where we learn $n$ individual projection heads $F^{proj.}_{\theta i}$, $i=\{1, \cdots, n\}$), while reusing $F^b_{\theta}$, $F^{prior}_{\theta}$ and $F^{head}_{\theta}$ trained in Stage \uppercase\expandafter{\romannumeral1}.

\subsection{Stage \uppercase\expandafter{\romannumeral1}: Diversified Segmentation}
In the conventional Probabilistic U-Net model \cite{kohl2018probabilistic}, there are two separate parts to achieve diversified segmentation. In the first part, the input $X$ alone is first used to generate the prior distribution $\mathbb{D}^{prior}(X)$, which is set as a multivariate normal distribution with a diagonal covariance matrix, denoted as $\mathcal{N}(\mu^{prior}, \sigma^{prior})$. The posterior distribution $\mathbb{D}^{post.}(X, A_{set}) = \mathcal{N}(\mu^{post.}, \sigma^{post.})$ is defined as the joint distribution of $X$ and all annotations $A_{set}$. The corresponding mean values and variances are obtained as
\begin{equation}
\begin{aligned}
\mu^{prior}, \sigma^{prior} = F_{\theta}^{prior}(X), \\
\mu^{post.}, \sigma^{post.} = F_{\theta}^{post.}(X,A_{set}),
\end{aligned}
\end{equation}
where $F_{\theta}^{prior}$ and $F_{\theta}^{post.}$ are two learnable neural encoders as shown in Fig.~\ref{fig_pipeline} (Left-top).
Here, $X$ and $A_{set}$ are concatenated together along the channel axis as the inputs of $F_{\theta}^{post.}$. A common Kullback–Leibler divergence (KL) loss is used to align the two distributions: 
\begin{equation}
\mathcal{L}_{kl} = KL(\mathbb{D}^{prior}(X), \mathbb{D}^{post.}(X, A_{set})).
\end{equation}

Then, the second part is for the target generation. Given a posterior distribution $\mathbb{D}^{post.}$, we randomly sample a latent code $z^{post.}_{random}\in\mathbb{R}^{D \times 1 \times 1}$, which is further scaled to $\hat{z}^{post.}_{random} \in \mathbb{R}^{D \times H \times W}$ as
the image size. Here, $D$ is the dimension of all latent codes. We then concatenate the scaled latent code and U-Net features together and feed them into a prediction head $F_{\theta}^{head}$ to generate the segmentation result:
\begin{equation}
\begin{aligned}
P^{post.}=F_{\theta}^{head}(\hat{z}^{post.}_{random},F_{\theta}^b(X)).
\end{aligned}
\end{equation}

Since we do not have the correspondence between $\hat{z}^{post.}_{random}$ and the label, we randomly select an annotation $A_{random}$ from $A_{set}$ as \cite{jungo2018effect,jensen2019improving}, and supervise the model training with a segmentation loss (\ie, Dice loss as \cite{schmidt2023probabilistic}):
\begin{equation}
\mathcal{L}^{rand.}_{seg} = DSC(P^{post.}, A_{random}).
\end{equation}

Following \cite{schmidt2023probabilistic}, we exploit a re-parameterization trick to sample $z^{post.}_{random}$ from the posterior distribution $\mathbb{D}^{post.}$ so as to back-propagate the gradients for the model training.

Nevertheless, the prediction diversity of such a conventional design is limited \cite{kohl2019hierarchical,baumgartner2019phiseg}. To address this, inspired by~\cite{zheng2019pluralistic} that leverages the prior distribution with a complementary task to improve the diversity of image inpainting, we propose a bound prediction task as the complement and also involve the prior distribution into the segmentation training. Specifically, the intersection and union (\ie, $A_{bound} = \{A_{inter.}, A_{union}\}$) of the annotation set are calculated as supervision labels. Then, by sampling $K$ latent codes from the prior distribution $\mathbb{D}^{prior}$, we can obtain $K$ segmentation results and their intersection and union as $P^{prior}_{bound} = \{P^{prior}_{inter.}, P^{prior}_{union}\}$, based on which we can compute the complementary segmentation loss:
\begin{equation}
\mathcal{L}_{bound} = DSC(P^{prior}_{inter.}, A_{inter.}) + DSC(P^{prior}_{union}, A_{union}).
\end{equation}

Overall, the model in Stage \uppercase\expandafter{\romannumeral1} is trained by a weighted sum of $\mathcal{L}_{kl}$, $\mathcal{L}^{rand.}_{seg}$ and $\mathcal{L}_{bound}$ as:
\begin{equation}\label{eq:stage1}
\mathcal{L}_{stage1} = \mathcal{L}_{kl} + \alpha \times \mathcal{L}^{rand.}_{seg} + \beta \times \mathcal{L}_{bound},
\end{equation}
where $\alpha$ and $\beta$ are two hyperparameters to balance the three losses. Such a design has several advantages: (1) we relax the predictions in uncertain regions to facilitate more diversified results; (2) the prior distribution is being trained anyway and we do not introduce any new model parameters; (3) the design is flexible and can be easily combined with other advanced model architectures \cite{baumgartner2019phiseg,kohl2019hierarchical}.

\subsection{Stage \uppercase\expandafter{\romannumeral2}: Personalized Segmentation}
Once Stage \uppercase\expandafter{\romannumeral1} learns a common latent space that can produce diversified and plausible segmentation predictions, the purpose of Stage \uppercase\expandafter{\romannumeral2} is to find specific expert prompts in the latent space that correspond to the individual expert annotators. As shown in Fig.~\ref{fig_pipeline} (Right), our idea is to employ $n$ individual projection heads to generate $n$ expert prompts from the U-Net features $F_{\theta}^b(X)$ as
\begin{equation}
z^{exp.}_i = Pooling[F_{\theta i}^{proj.}(F_{\theta}^b(X)], \, i = 1,\cdots,n
\end{equation}
where $z^{exp.}_i \in \mathbb{R}^{D \times 1 \times 1}$ denotes the specific latent code relevant for the $i$-th expert. It is obtained via a global average pooling operation for the 2D outputs of the $i$-th projection head $F_{\theta i}^{proj.}$, to capture global annotation characteristics.

Then, to ensure all generated expert prompts are in the learned common latent space, we further introduce a cross-attention operation to map $z^{exp.}_i$ into $\Tilde{z}^{exp.}_i$. In particular, we first randomly sample $M$ latent codes from the fixed prior distribution $D^{prior}$ as a prior bank $z^{prior}_{bank} \in \mathbb{R}^{D \times 1 \times M}$. We then use $z^{exp.}_i$ as the query and $z^{prior}_{bank}$ as the key and value for cross-attention:
\begin{equation}
\begin{aligned}
\Tilde{z}^{exp.}_i = Softmax({z^{exp.}_i}^{\top} \cdot z^{prior}_{bank})\cdot {z^{prior}_{bank}}^{\top}.
\end{aligned}
\end{equation}

In this way, $\Tilde{z}^{exp.}_i$ is constrained into the frozen prior distribution $D^{prior}(X)$, with respect to the input $X$. After that, we use the fixed $F_{\theta}^{head}$ to transform the expert prompt into the final segmentation result $P^{exp.}_i$, which is used to compare with the corresponding expert annotation $A_i$ for computing the segmentation loss as
\begin{equation}
\begin{aligned}
\mathcal{L}_{seg}^{corr.} &= \sum^{n}_{i=1} DSC(P^{exp.}_i, A_i) \\
s.t., P^{exp.}_i &= F_{\theta}^{head}(\hat{z}^{exp.}_{i},F_{\theta}^b(X)),
\end{aligned}
\end{equation}
where $\hat{z}^{exp.}_{i}$ is the scaled version of $\Tilde{z}^{exp.}_{i}$. Note that, $\mathcal{L}_{seg}^{corr.}$ is only used to train the $n$ projection heads, while other modules are fixed in Stage \uppercase\expandafter{\romannumeral2}. This design principle of our Stage \uppercase\expandafter{\romannumeral2} personalized segmentation part is general and can be extended into other powerful generative methods \cite{rahman2023ambiguous}.

\section{Experiments and Results}
\label{sec:results}
\input{tables/tab_npc}
\subsection{Dataset}
\label{dataset}
We evaluate our method on the two segmentation datasets: our in-house dataset (NPC-170) and the public lung nodule segmentation dataset (LIDC-IDRI) \cite{armato2011lung}.

The NPC-170 dataset contains 170 subjects with Nasopharyngeal Carcinoma in three Magnetic Resonance Imaging (MRI) contrasts (T1, T1-Contrast, and T2).
For each sample, four senior radiologists from different locations (with experiences of around 5-10 years) individually annotated the GTVp of Nasopharyngeal Carcinoma. We divide the 170 subjects into separate training, validation, and testing sets (\ie, 100, 20, and 50 subjects, respectively). 2D medical image segmentation is employed in the experiments and, in total, there are 6134, 1126, and 3058 slices for the model training, validation, and testing, respectively.

The LIDC-IDRI dataset contains 1609 2D thoracic CT scans belonging to 214 subjects. Each scan is equipped with four binary masks to indicate the lung nodule. Note that, 12 radiologists participated in the annotation process \cite{armato2011lung}. To simulate the expert preferences, we manually rank the areas of four provided annotations, as the setting in \cite{HumanError2020}. In this way, the first expert is supposed to be conservative in the sense of only annotating the minimum regions while the fourth expert prefers to annotate any suspicious lung nodule area, \ie, producing the largest annotation regions. Following \cite{wang2023medical}, a four-fold cross-validation setting at the patient level is used in the experiments.

\subsection{Implementation Details}
The inputs from both the NPC-170 and LIDC-IDRI datasets are center-cropped into a fixed size of $128\times128$. Then, for the samples on NPC-170, we normalize them into zero mean and unit variance. Random flips, rotation, or adding random noise are applied to augment the training data. On the LIDC-IDRI dataset, we follow \cite{wang2023medical} to perform the data pre-processing and use the common flip and rotation operations to augment the samples on the LIDC-IDRI dataset.

For the model training, we adopt the Adam optimizer with an initial learning rate of 1e-4. The loss weight $\alpha$ and the distribution dimension $D$ are set as 1 and 6, as \cite{schmidt2023probabilistic}. The overall training epochs are 300, including 100 epochs for Stage \uppercase\expandafter{\romannumeral1} and 200 epochs for Stage \uppercase\expandafter{\romannumeral2}. The size of $z_{bank}^{prior}$, \ie, $M$ is set as 100. $K$ and $\beta$ are set as 10 and 0.5, which will be discussed in Section~\ref{sec:dis_hyperpara}. An L2-regularization term is also used to avoid over-fitting as \cite{kohl2018probabilistic,schmidt2023probabilistic}. All experiments are conducted in an identical environment with a single NVIDIA GeForce RTX 3090 GPU. The overall number of parameters is 28.46 M and the projection head is relatively lightweight (\ie, 22.03 K for each head).

\subsection{Evaluation Metrics}
Considering our D-Persona model consists of two stages: diversified and personalized segmentation, we thus evaluate them separately.  First, we use two common set-to-set metrics for the diversity estimation, \ie, the Generalized Energy Distance $GED$ \cite{bellemare2017cramer,kohl2018probabilistic} and the soft Dice score $Dice_{soft}$ \cite{wang2023medical,ji2021learning}. The former is used to measure the prediction diversity and the latter indicates the reliability of different generated results. Specifically, $GED$ is defined as
\begin{equation}
\begin{aligned}
GED = 2\mathbb{E}[d(P, A)] - \mathbb{E}[d(P, P')] - \mathbb{E}[d(A, A')],  
\end{aligned}
\end{equation}
where $P$, $P'$ and $A$, $A'$ are independent samples from the prediction set $P_{set}$ and annotation set $A_{set}$, respectively, and $d$ denotes the distance function $d(a,b) = 1- IoU(a,b)$ as in \cite{kohl2018probabilistic}. In general, a lower $GED$ indicates greater dispersion and variability in the segmentation results.
Moreover, $Dice_{soft}$ receives the soft predictions $P_{soft}$ and soft annotations $A_{soft}$ (\ie, the average of multiple predictions and annotations) and uses varied thresholds to conduct $T$ times of binary evaluations as:
\begin{equation}
\begin{aligned}
Dice_{soft} = \frac{1}{T}\sum^{T}_{i=1} Dice([P_{soft} > \tau_i], [A_{soft} > \tau_i])
\end{aligned}
\end{equation}
where $\tau$ is a threshold selected from the set \{0.1, 0.3, 0.5, 0.7, 0.9\} with $T = 5$.

\begin{figure}[htp]
  \centering
   \includegraphics[width=0.8\linewidth]{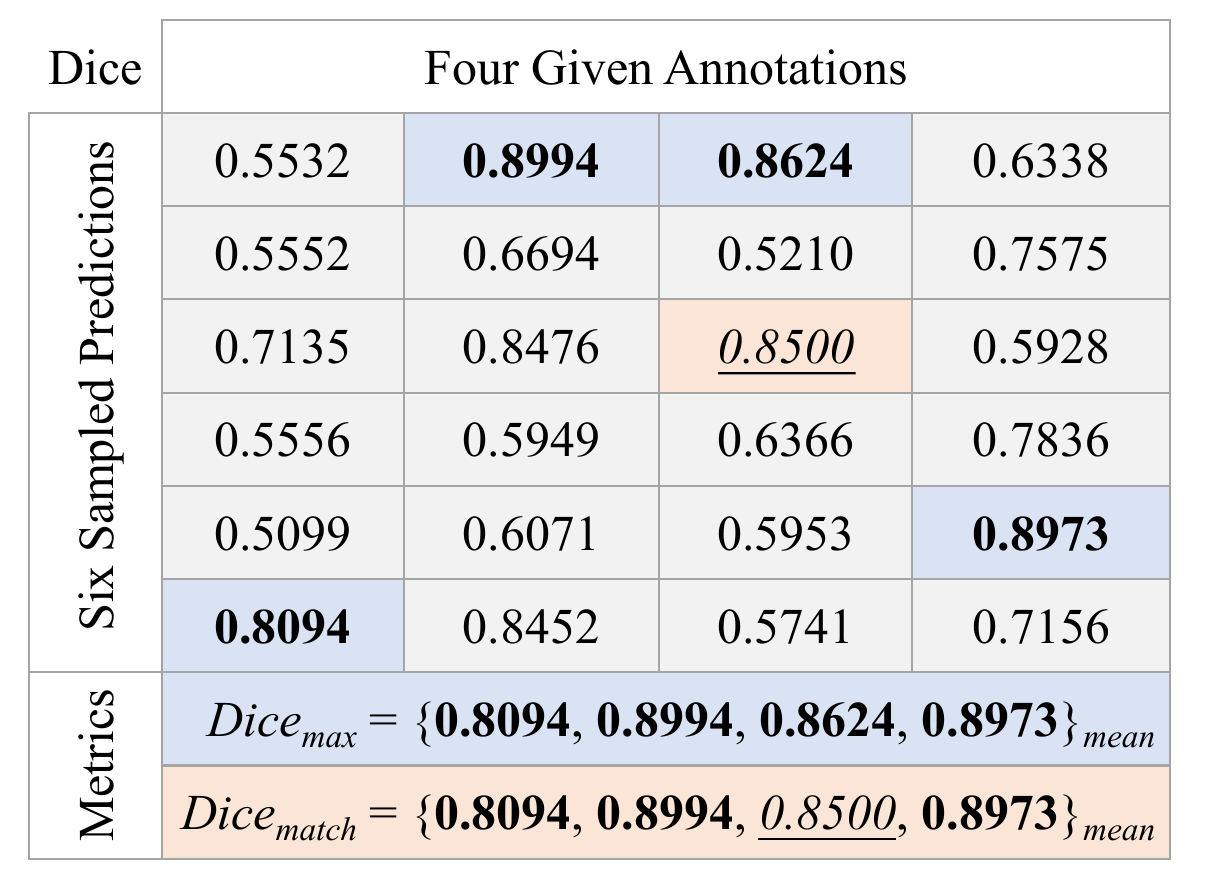}
   \caption{Exemplar explanation of $Dice_{max}$ and $Dice_{match}$ in a given $4 \times 6$ Dice matrix. $Dice_{max}$ averages the maximum scores of individual columns and $Dice_{match}$ further constrains a one-to-one matching between the prediction and annotation sets.}
   \label{fig_metric}
\end{figure}
Second, we further use the set-to-set metrics $Dice_{max}$ and $Dice_{match}$ to evaluate the personalized segmentation in our framework. As Fig.~\ref{fig_metric} shows, $Dice_{max}$ quantifies the optimal overlap between $P_{set}$ and $A_{set}$. $Dice_{match}$ further imposes a stringent one-to-one matching criterion between the sets. In other words, $Dice_{max}$ is an upper bound of $Dice_{match}$ and their differences serve as an indicator to determine whether each expert annotation corresponds with a unique and closely aligned prediction. 

Additionally, we also provide each annotator a Dice score (denoted as $Dice_{A(i)}$, $i=1,\cdots,n$),  to reflect the personalized segmentation performance for the respective human expert. Then, $Dice_{mean}$ further gives the average personalized performance of the $n$ experts.

\input{tables/tab_lidc}
\subsection{Performance on NPC-170}
\label{sec:performance}

Table~\ref{tab_npc} gives the quantitative performance of our proposed D-Persona in comparison with other existing works on the in-house NPC-170 dataset, indicating several findings: 
(1) the individually trained U-Net models, where only annotations from a single expert are used,
achieve superior results for the targeted expert but significantly poor performance for the others, highlighting the great inter-observer variability; 
(2) for generation-based methods, our Stage \uppercase\expandafter{\romannumeral1} exhibits remarkable segmentation diversity than the baseline model, as evidenced by leading scores in $GED$ and $Dice_{soft}$; 
(3) we also observe an increase in personalized bounds in Stage \uppercase\expandafter{\romannumeral1} (\eg, $Dice_{max}$ improves from 74.18\% of the baseline to 76.54\% of our model), and further improvements can be achieved by increasing the number of sampled latent codes (\eg, a 1.16\% gain in $Dice_{max}$ with 50 random samples); 
(4) in Stage \uppercase\expandafter{\romannumeral2}, our two-stage personalization strategy surpasses the performance of other one-stage methods (\eg, a 1.10\% gain in $Dice_{mean}$ than the second-best model \cite{schmidt2023probabilistic}), demonstrating the superiority of our approach. 
Note that, when compared with the individually trained models, our D-Persona Stage \uppercase\expandafter{\romannumeral2} does not perform better for the personalized results specific to the particular annotator except $A_4$. This also happens to all the compared one-stage methods.

\subsection{Performance on LIDC-IDRI}
Table~\ref{tab_lidc} extends the evaluation of our model to the LIDC-IDRI dataset, where the average performance of a four-fold cross-validation is given. The individually trained results indicate that the inter-observer variance is less pronounced compared to our in-house NPC-170 dataset. Then, echoing the findings from NPC-170, our D-Persona model consistently outperforms the baseline Probabilistic U-Net \cite{kohl2018probabilistic} in achieving superior diversified segmentation performance. Furthermore, the second stage of our proposed D-Persona model exhibits better performance in both the set-to-set evaluations and the one-to-one comparisons. Meanwhile, in Stage \uppercase\expandafter{\romannumeral2}, we can see that the average personalized performance has reached the upper bound (\ie, both $Dice_{match}$ and $Dice_{mean}$: 89.17\%) on LIDC-IDRI, highlighting its effectiveness in personalized medical image segmentation.

\section{Discussions}
\subsection{Visual Results}
\begin{figure}[htbp]
  \centering
   \includegraphics[width=1\linewidth]{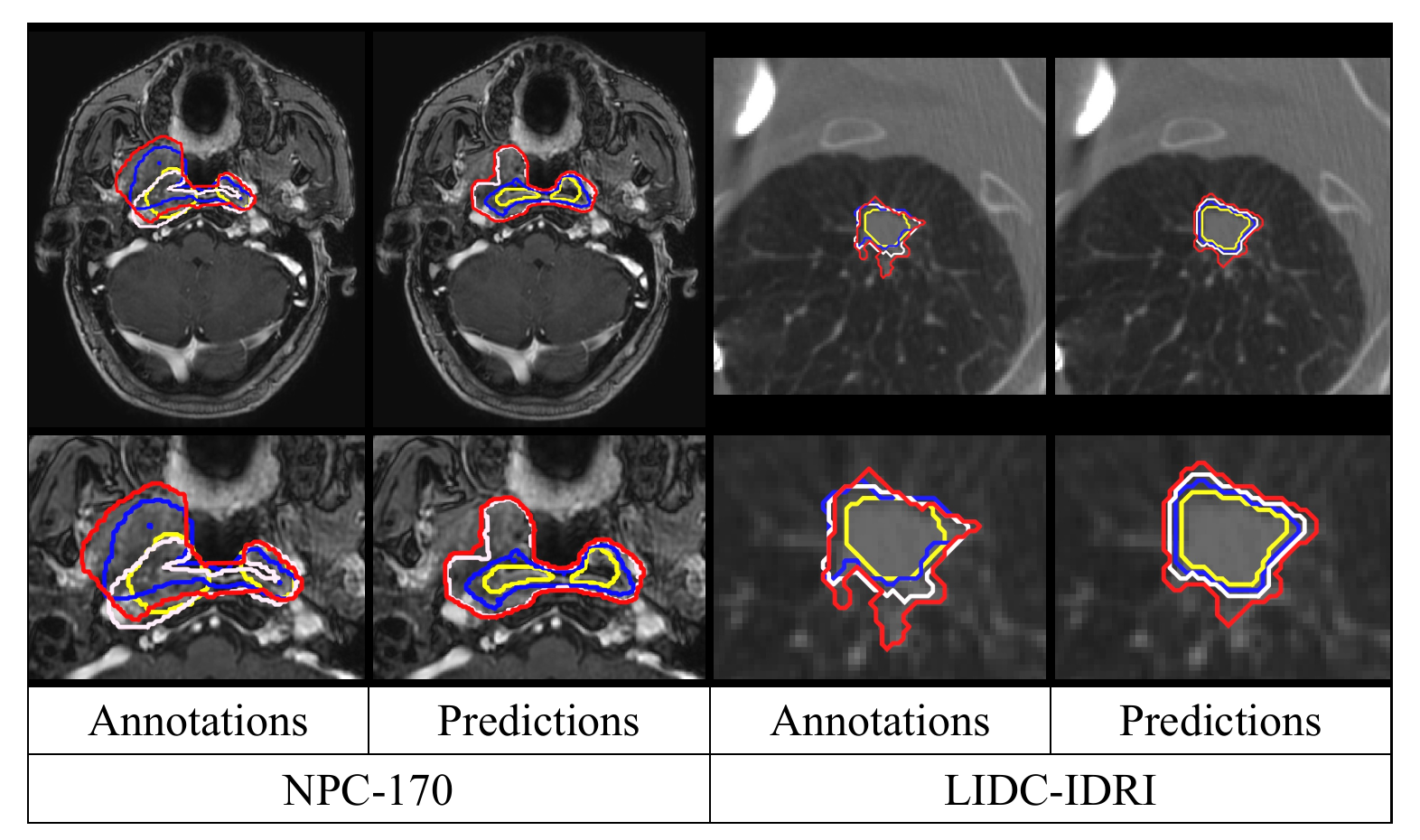}
   \caption{Diversified segmentation results of Stage \uppercase\expandafter{\romannumeral1} in our proposed D-Persona framework on the NPC-170 (Left) and LIDC-IDRI (Right) datasets. Different colors denote different delineations, which shows that our model can generate diverse and plausible predictions.}
   \label{fig_stage1}
\end{figure}

Fig.~\ref{fig_stage1} shows the diversified segmentation results of Stage \uppercase\expandafter{\romannumeral1} in our D-Persona model. We can see that, on the two datasets, our model can generate varying and plausible predictions for the target regions, which brings potential benefits for providing different opinions in clinical scenarios.
\vspace{-0.5cm}
\input{tables/ablation_para}

\begin{figure*}[htbp]
  \centering
   \includegraphics[width=1\linewidth]{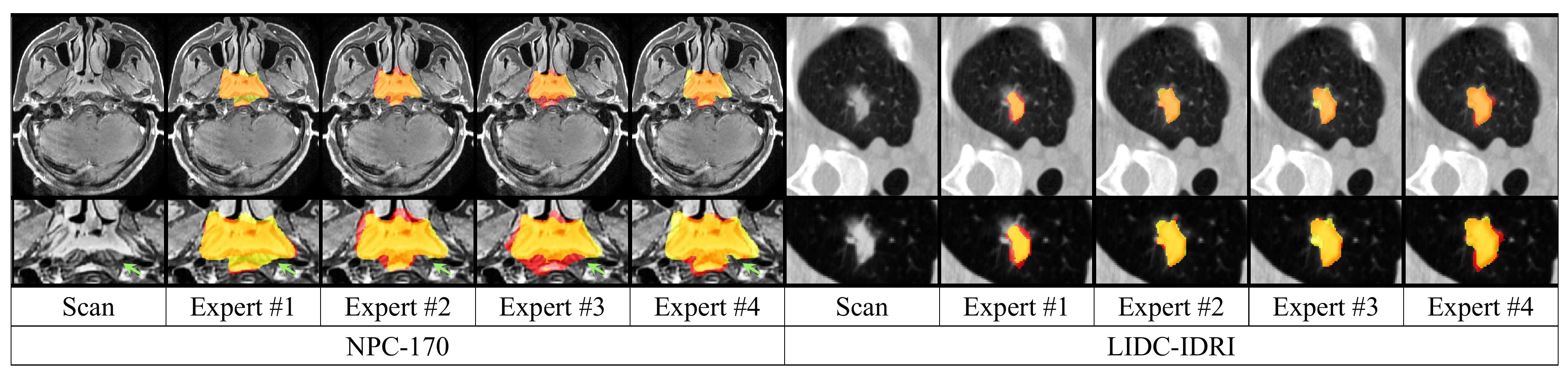}
   \caption{Personalized segmentation results of Stage \uppercase\expandafter{\romannumeral2} in our proposed D-Persona framework on the NPC-170 (Left) and LIDC-IDRI (Right) datasets. Compared to expert annotations (Red), our model can generate corresponding personalized segmentation results (Yellow). Particularly, our model successfully captures the underlying annotation preferences, \ie, from conservative to aggressive styles, as shown in the four LIDC-IDRI results on the right.}
   \label{fig_stage2}
\end{figure*}

Fig.~\ref{fig_stage2} further illustrates the personalized segmentation results of Stage \uppercase\expandafter{\romannumeral2} in the proposed model. First, on the NPC-170 dataset, our D-Persona can detect most of the targets, producing predictions in complex anatomical structures as the given expert annotations, indicated by the green arrows in Fig.~\ref{fig_stage2}. Second, on LIDC-IDRI, as we simulated the expert preferences from conservative to aggressive styles (see Section~\ref{dataset}), the right part of Fig.~\ref{fig_stage2} reveals that our model can successfully capture the underlying preferences, demonstrating the superiority of our proposed D-Persona model for personalized segmentation.

\label{sec:discussions}
\subsection{Selection of Hyper-parameters}
\label{sec:dis_hyperpara}
To control the diversity of generated segmentation results, we here discuss the selection of hyper-parameters $K$ and $\beta$ in Stage \uppercase\expandafter{\romannumeral1} of our model. 
$K$ is the size of the sampled latent codes in the prior distribution $\mathbb{D}^{prior}$. Since we applied the bound-constrained loss to supervise the learning of $\mathbb{D}^{prior}$ and a smaller $K$ could make the segmentation results approach the fixed annotation bounds. The top of Table~\ref{tab_parameters} shows that, on the NPC-170 dataset, setting $K$ as 10 can achieve better diversified performance for the GTVp delineation of Nasopharyngeal Carcinoma.

At the same time, $\beta$ is used to balance the proposed bound loss with other losses in Equation \eqref{eq:stage1}. The bottom results of Table~\ref{tab_parameters} indicate that selecting $\beta$ as 0.5 for the model training achieves the highest $Dice_{soft}$ and second-best $GED$ values on the NPC-170 dataset. We thus set $\beta$ as 0.5 for all experiments in this paper.

\subsection{Comparisons with Crowdsourcing Methods}
\input{tables/ablation_voting}
As aforementioned, crowdsourcing methods aim to combine multiple annotations for the model training. Table~\ref{tab_voting} shows the performance of several conventional crowdsourcing methods, including majority voting (MV), Random Selection (RS) \cite{jungo2018effect} and STAPLE \cite{warfield2004simultaneous} on the NPC-170 dataset.
We can see that, our proposed D-Persona model achieves the highest performance in $Dice_{mean}$ than all heuristic crowdsourcing works (\eg, a 0.95\% gain than the second-best model \cite{jungo2018effect}). Furthermore, our model can provide diversified segmentation results at the same time, facilitating more clinical usage through a unified model.

\subsection{Limitation and Future Work}
As mentioned in Section~\ref{sec:performance}, on the NPC-170 dataset, it remains difficult for our model to surpass the individually trained U-Net models for personalized segmentation. This suggests that preserving the distinctive preference of a particular expert is still a challenge in this task, especially when the training labels vary significantly. In other words, improving the predictions having an individual expert's style by leveraging different annotations from other experts is not fully resolved, which is worth future investigation.
\section{Conclusion}
\label{sec:conclusion}
In this paper, we have presented a two-stage framework (D-Persona) for the multi-rater medical image segmentation problem, pointing out a new perspective that diversified and personalized segmentation can be addressed at the same time.
Our main idea is to conduct diversified segmentation first and then query personalized results from the diversified ones. 
Specifically, an effective bound-constrained loss was proposed to improve the segmentation diversity, and a common latent space was constructed in Stage \uppercase\expandafter{\romannumeral1} to capture the annotation variability.
Then, in the second stage, an attention-based projection mechanism was used to find specific expert prompts from the common latent space, so as to perform personalized segmentation. Extensive experiments on the two datasets have demonstrated that our proposed D-Persona model significantly outperforms other existing works in multi-rater medical image segmentation.

\noindent\textbf{Societal Impacts.} Our proposed model was trained and evaluated on the two available but limited datasets and may be impacted by the dataset bias \cite{ricci2022addressing}, which might result in unconvincing predictions in practical applications.

{
    \small
    \bibliographystyle{ieeenat_fullname}
    \bibliography{main}
}

\end{document}

%% file: preamble.tex
\usepackage[dvipsnames]{xcolor}


%% file: tables/tab_npc.tex
\begin{table*}[htp]
\caption{Performance of our proposed D-Persona framework and several public methods for multi-rater medical image segmentation on our in-house NPC-170 dataset. We show the results of individually trained U-Net models (\ie, using annotations from a single expert, Top), generation-based methods (with different sampling numbers, Middle), and personalized segmentation works (Bottom), respectively.}
\centering
\resizebox{1.0\linewidth}{!}{
\begin{tabular}{c|cc|cc|ccccc}
\toprule[1.5pt]
\multirow{2}{*}{Method} & \multicolumn{2}{c|}{Diversity Performance} & \multicolumn{2}{c|}{Personalized Bounds (\%)}   & \multicolumn{5}{c}{Personalized Segmentation Performance (\%)} \\

\cline{2-10} & $GED$ $\downarrow$ & $Dice_{soft}\uparrow$ (\%) & $Dice_{max}\uparrow$ &   $Dice_{match}\uparrow$  &  $Dice_{A1}\uparrow$&  $Dice_{A2}\uparrow$ &  $Dice_{A3}\uparrow$ &  $Dice_{A4}\uparrow$ &  $Dice_{mean}\uparrow$        \\ \midrule
U-Net ($A_1$) &0.4888 &73.69 &\multicolumn{2}{c|}{\multirow{4}{*}{N/A}} &\textbf{80.90} &70.51 &67.14 &72.57 &72.78 \\
U-Net ($A_2$) &0.5086 &73.64 & & &72.05 &\textbf{75.62} &69.93 &71.85 &72.36 \\
U-Net ($A_3$) &0.5077 &72.37 & & &69.46 &73.09 &\textbf{76.25} &69.86 &72.17 \\
U-Net ($A_4$) &0.5433 &71.63 & & &73.71 &68.77 &66.08 &\textbf{73.54} & 70.52\\ \midrule

Prob. U-Net \cite{kohl2018probabilistic} (\#10) &0.4478 &75.33 &74.18 &74.17 & \multicolumn{5}{c}{\multirow{6}{*}{N/A}}  \\
Prob. U-Net \cite{kohl2018probabilistic} (\#30) &0.4466 &75.34 &74.23 &74.22 & & & & & \\
Prob. U-Net \cite{kohl2018probabilistic} (\#50) &0.4465 &75.34 &74.24 &74.24 & & & & & \\
D-Persona (Stage \uppercase\expandafter{\romannumeral1}, \#10) &0.2553 &79.99 &76.54 &76.40 & & & & & \\
D-Persona (Stage \uppercase\expandafter{\romannumeral1}, \#30) &0.2385 &80.40 &77.50 &77.43 & & & & & \\
D-Persona (Stage \uppercase\expandafter{\romannumeral1}, \#50) &\textbf{0.2359} &\textbf{80.43} &\textbf{77.70} &\textbf{77.67} & & & & & \\ \midrule

CM-Global \cite{tanno2019learning} &0.4375 &76.28 &75.00 &75.00 &77.92 &\textbf{74.65} &71.97 &\textbf{75.47} &75.00 \\
CM-Pixel \cite{HumanError2020} &0.4400 &76.39 &74.92 &74.92 &78.97 &73.93 &71.65 &75.12 &74.92 \\
TAB \cite{liao2023transformer} &\textbf{0.2760} &78.90 &76.68 &75.15 &77.33 &73.45 &74.83 &71.67 &74.32 \\
Pionono \cite{schmidt2023probabilistic}  &0.3924 &76.96 &75.58 &75.44 &78.87 &74.11 &71.97 &75.41 &75.09 \\
D-Persona (Stage \uppercase\expandafter{\romannumeral2}) &0.3199 &\textbf{79.01} &\textbf{77.26} &\textbf{76.61} &\textbf{79.78} &74.60 &\textbf{75.22} & 75.17 &\textbf{76.19}  \\
\bottomrule[1.5pt]
\end{tabular}}
\label{tab_npc}
\end{table*}

%% file: tables/tab_lidc.tex
\begin{table*}[htp]
\caption{Performance of our proposed D-Persona framework and several public methods for multi-rater medical image segmentation on the public LIDC-IDRI dataset. We show the results of individually trained U-Net models (\ie, using annotations from a single expert, Top), generation-based methods (with different sampling numbers, Middle), and personalized segmentation works (Bottom), respectively.}
\centering
\resizebox{1.0\linewidth}{!}{
\begin{tabular}{c|cc|cc|ccccc}
\toprule[1.5pt]
\multirow{2}{*}{Method} & \multicolumn{2}{c|}{Diversity Performance} & \multicolumn{2}{c|}{Personalized Bounds (\%)}   & \multicolumn{5}{c}{Personalized Segmentation Performance (\%)} \\

\cline{2-10} & $GED$ $\downarrow$ & $Dice_{soft}\uparrow$ (\%) & $Dice_{max}\uparrow$ &   $Dice_{match}\uparrow$  &  $Dice_{A1}\uparrow$&  $Dice_{A2}\uparrow$ &  $Dice_{A3}\uparrow$ &  $Dice_{A4}\uparrow$ &  $Dice_{mean}\uparrow$        \\ \midrule
U-Net ($A_1$) &0.3062 &86.59 &\multicolumn{2}{c|}{\multirow{4}{*}{N/A}} &\textbf{87.80} &87.47 &85.49 &80.67 &85.36 \\
U-Net ($A_2$) &0.2459 &88.43 & & &87.16 &\textbf{89.08} &88.59 &85.15 &87.50\\
U-Net ($A_3$) &0.2436 &88.20 & & &85.29 &88.48 &\textbf{89.40} &87.20 &87.59\\
U-Net ($A_4$) &0.2962 &85.83 & & &80.80 &85.48 &88.22 &\textbf{88.90} &85.85\\ \midrule

Prob. U-Net \cite{kohl2018probabilistic} (\#10) &0.2181 &88.79 &88.60 &88.43 & \multicolumn{5}{c}{\multirow{6}{*}{N/A}}  \\
Prob. U-Net \cite{kohl2018probabilistic} (\#30) &0.2169 &88.79 &88.80 &88.73 & & & & & \\
Prob. U-Net \cite{kohl2018probabilistic} (\#50) &0.2168 &88.80 &88.87 &88.81 & & & & & \\
D-Persona (Stage \uppercase\expandafter{\romannumeral1}, \#10) &0.1461 &90.24 &90.75 &90.51 & & & & & \\
D-Persona (Stage \uppercase\expandafter{\romannumeral1}, \#30) &0.1375 &90.42 &91.23 &91.16 & & & & & \\
D-Persona (Stage \uppercase\expandafter{\romannumeral1}, \#50) &\textbf{0.1358} &\textbf{90.45} &\textbf{91.37} &\textbf{91.33} & & & & & \\ \midrule

CM-Global \cite{tanno2019learning} &0.2432 &88.53 &87.51 &87.51 &86.13 &88.76 &88.99 &86.18 &87.51 \\
CM-Pixel \cite{HumanError2020} &0.2407 &88.64 &87.72 &87.72 &85.99 &88.81 &89.31 &86.77 &87.72 \\
TAB \cite{liao2023transformer} &0.2322 &86.35 &87.11 &86.08 &85.00 &86.35 &86.77 &85.77 &85.97 \\
Pionono \cite{schmidt2023probabilistic}  &0.1502 &90.00 &90.10 &88.97 &87.94 &89.11 &89.55 &\textbf{88.76} &88.84 \\
D-Persona (Stage \uppercase\expandafter{\romannumeral2}) &\textbf{0.1444} &\textbf{90.31} &\textbf{90.38}&\textbf{89.17}&\textbf{88.54}&\textbf{89.50}&\textbf{90.03} &88.60 &\textbf{89.17}  \\
\bottomrule[1.5pt]
\end{tabular}}
\label{tab_lidc}
\end{table*}

%% file: tables/ablation_para.tex
\begin{table}[htp]
\caption{Ablation studies of different $K$ and $\beta$ in Stage \uppercase\expandafter{\romannumeral1} of our proposed D-Persona framework on the NPC-170 dataset. The sampling number is set as 50 for all comparisons.}
\label{tab_parameters}
\begin{subtable}{0.5\linewidth}
\centering
\resizebox{\linewidth}{!}
{
\begin{tabular}{c|c|cc}
\toprule[1.5pt]
$K$  & $\beta$  & $GED\downarrow$ & $Dice_{soft}\uparrow$(\%) \\ 
\midrule
6 & \multirow{4}{*}{\, 0.5 \,}  &0.2545   &78.81  \\
8 &  &0.2512  &79.48  \\
10 & &\textbf{0.2359}  &\textbf{80.43}   \\
12 & &0.2389  &80.17  \\
14 & &0.2368  &80.08   \\
\bottomrule[1.5pt]
\end{tabular}
}
\end{subtable}%
\begin{subtable}{0.5\linewidth}
\centering
\resizebox{\linewidth}{!}
{
\begin{tabular}{c|c|cc}
\toprule[1.5pt]
$K$   & $\beta$  & $GED\downarrow$ & $Dice_{soft}\uparrow$(\%) \\ 
\midrule
\multirow{4}{*}{10} & 0.01     &0.4378   &75.80    \\
                    & 0.1     &0.3358  &77.95   \\
                     &\, 0.5 \,     &0.2359  &\textbf{80.43}  \\
                     & 1   &\textbf{0.2170}  &80.20    \\ 
                      & 2          &0.2581  &77.73  \\
\bottomrule[1.5pt]
\end{tabular}
}
\end{subtable}%
\end{table}

%% file: tables/ablation_voting.tex
\begin{table}[htp]
\caption{Comparisons between several crowdsourcing methods including majority voting (MV), random selection (RS) \cite{jungo2018effect}, STAPLE \cite{warfield2004simultaneous} and our proposed D-Persona model on NPC-170.}
\centering
\resizebox{1.0\linewidth}{!}{
\begin{tabular}{c|ccccc}
\toprule[1.5pt]
Method&  $Dice_{A1}\uparrow$&  $Dice_{A2}\uparrow$ &  $Dice_{A3}\uparrow$ &  $Dice_{A4}\uparrow$ &  $Dice_{mean}\uparrow$        \\ \midrule
MV &76.80 &71.28 &70.24 &73.64 &72.99  \\
RS~\cite{jungo2018effect} &78.45 &74.92 &72.04 &\textbf{75.54} &75.24  \\
STAPLE~ \cite{warfield2004simultaneous} &75.75 &\textbf{75.76} &72.29 &74.07 &74.47  \\
D-Persona &\textbf{79.78} &74.60 &\textbf{75.22} & 75.17 &\textbf{76.19}  \\
\bottomrule[1.5pt]
\end{tabular}}
\label{tab_voting}
\end{table}